\documentclass[journal]{IEEEtran}
\usepackage{latexsym}

\usepackage{cite}

\ifCLASSINFOpdf
   \usepackage[pdftex]{graphicx}
  \graphicspath{Figures/}
  
\else
 \fi

\usepackage[cmex10]{amsmath}

\usepackage{algorithm}
\usepackage{algpseudocode}
\usepackage{array}
\usepackage{bm}   
\usepackage{multirow} 
\usepackage{bigdelim} 
\usepackage{caption} 
\usepackage{amssymb}
 
\usepackage{booktabs}
\usepackage{xcolor}
\usepackage{makecell}

\newcommand{\ck}{\checkmark}
\usepackage{xurl}
\usepackage{tabularx}

\ifCLASSOPTIONcompsoc
  \usepackage[caption=false,font=normalsize,labelfont=sf,textfont=sf]{subfig}
\else
  \usepackage[caption=false,font=footnotesize]{subfig}
\fi

\usepackage{stfloats}

\ifCLASSOPTIONcaptionsoff
  \usepackage[nomarkers]{endfloat}
 \let\MYoriglatexcaption\caption
 \renewcommand{\caption}[2][\relax]{\MYoriglatexcaption[#2]{#2}}
\fi

\begin{document}
%
\title{Is it Safe to Drive? An Overview of Factors, Challenges, and Datasets for Driveability Assessment in Autonomous Driving}
%
%
%

\author{Junyao~Guo, Unmesh Kurup, Mohak Shah
        
\thanks{The authors are with Advanced AI, LG Silicon Valley Lab, US
}
}

\maketitle


\begin{abstract}
With recent advances in learning algorithms and hardware development, autonomous cars have shown promise when operating in structured environments under good driving conditions. However, for complex, cluttered and unseen environments with high uncertainty, autonomous driving systems still frequently demonstrate erroneous or unexpected behaviors, that could lead to catastrophic outcomes. Autonomous vehicles should ideally adapt to driving conditions; while this can be achieved through multiple routes, it would be beneficial as a first step to be able to characterize \textit{Driveability} in some quantified form. To this end, this paper aims to create a framework for investigating different factors that can impact driveability. Also, one of the main mechanisms to adapt autonomous driving systems to any driving condition is to be able to learn and generalize from representative scenarios. The machine learning algorithms that currently do so learn predominantly in a supervised manner and consequently need sufficient data for robust and efficient learning. Therefore, we also perform 
a comparative overview of 45 public driving datasets that enable learning and publish this dataset index at \textit{https://sites.google.com/view/driveability-survey-datasets}. Specifically, we categorize the datasets according to use cases, and highlight the datasets that capture complicated and hazardous driving conditions which can be better used for training robust driving models. Furthermore, by discussions of what driving scenarios are not covered by existing public datasets and what driveability factors need more investigation and data acquisition, this paper aims to encourage both targeted dataset collection and the proposal of novel driveability metrics that enhance the robustness of autonomous cars in adverse environments.
\end{abstract}

%
\IEEEpeerreviewmaketitle
\section{Introduction}
Despite testing autonomous cars in highly controlled settings, these cars still occasionally fail in making correct decisions, often with catastrophic results\footnote{These apply to autonomous vehicles in general.}.
There have been several accidents reported recently \cite{tian2017deeptest} due to failure of the autonomous capability of these cars. According to the accident records, the failures are most likely to happen in complex or unseen driving environments. The fact remains that while autonomous cars can operate well in controlled or structured environments such as highways, they are still far from reliable when operating in cluttered, unstructured or unseen environments \cite{muoio6scenarios}. 

To adapt autonomous driving systems to all types of driving conditions, it would be beneficial to first characterize the \textit{Driveability} of a scene\footnote{We adopt the definition of a \textit{scene} proposed in \cite{ulbrich2015defining}; i.e., ``A scene describes a snapshot of the environment including the scenery and dynamic elements, as well as all actors’ and observers’ self-representations, and the relationships among those entities."}. This can then lead to addressing two core issues of autonomous driving or advanced driver-assistance systems (ADAS): 1) policy learning for driver control hand-off; and 2) incorporation of \textit{driveability} in the autonomous vehicle's decision making, planning and testing. These two different application fields also suggest that \textit{driveability} could be quantified in different forms, either as a single metric or a composition of metrics. For example, with ADAS and current Level 2 or 3 autonomy, a scene can be simply defined as \textit{driveable} if the car can operate safely in autonomous mode. When a non-driveable scene is detected, the autonomous car can hand over control to the human driver in a timely manner \cite{hecker2018failure}. However, in the long term where Level 4 or 5 autonomy is targeted, it is not possible to hand over control to the driver. This restriction means that a richer representation of driveability is needed; one that is informative enough for the car to take proactive measures to prevent catastrophic failure. For instance, an autonomous car can selectively use sensors in scenes with high driveability whereas request further support from cloud computing for detailed analysis of less-driveable scenes \cite{kumar2012cloud}. In scenes with particularly low driveability, contingency plans might need to be executed such as deceleration or even making a full stop. The driveability information could also be useful when processed offline. For example, it can be used to build a safety map that not only guides road users to plan alternate routes that are safer, but also identifies unsafe areas that need repair \cite{cafiso2011safety}. Such information can also be used by insurance companies to quantify operating risk \cite{fuchs2016risk, hsu2017subjective}. 

The concept of driveability is not new. It has been proposed for measuring road conditions \cite{tsukada2007evaluation} and modeling human driver performance \cite{bekiaris2003drivability}. However, there exists no unified concept of driveability in the domain of autonomous driving. Related usages of driveability include ``driveability map" which is a map divided into cells categorized as driveable or non-driveable for motion planning \cite{leonard2008perception, thrun2006stanley, sivaraman2014dynamic, 7823116}, and ``object driveability" which determines if an object can be driven over without causing damage to the vehicle \cite{ferguson2018determining}. The concept that is most similar to that considered in this paper is ``scene driveability" proposed in two recent studies \cite{hecker2018failure, scheel2018situation}. Scene driveability measures how easy a scene is for an autonomous car to make full decisions on steering angles and accelerations or to perform a specific task such as lane changing. However, none of these studies reason about what makes a scene driveable or non-driveable at a high level in the context of autonomous driving. 

While it may not be difficult for human drivers to tell whether an environment is safe, it is far from obvious for an autonomous car because the driveability is affected by a variety of factors. Therefore, we first provide one taxonomy of those factors including both environmental conditions and road user behaviors, and particularly review representative approaches that handle the risks originating from each factor. Then, established performance metrics for driveability assessment are summarized. The development of both robust methods and novel metrics depends, in part, on having access to large-scale naturalistic and diverse driving datasets. This will have consequence not just in understanding of driveability conditions and developing novel quantification metrics, but will also assist subsequent validation and verification as well as incorporation of driveability aspects in autonomous driving policy. Therefore, as a major contribution of this paper, an exhaustive and comparative study of publicly available datasets for autonomous driving research is presented. To provide practical guidance on when to use which dataset, we categorize the datasets according to autonomous driving tasks. More importantly, we highlight the datasets that capture low-driveability scenes using which models could be trained to better handle traffic hazards and risks. 

By reviewing existing literature and datasets from a driveability perspective, this paper accomplishes the following:
\begin{itemize}
\item Identifies both environmental and behavioral factors that contribute to driveability of the scene and sheds light on the limitations of existing approaches in handling low-driveability scenarios.
\item Reviews established metrics for driveability assessment, identifies their limitations and proposes the need of principled driveability metrics for enhancing reliability of autonomous driving systems.
\item Provides a practical reference for the autonomous driving research community of 45 publicly available driving datasets, highlighting large-scale datasets that are suitable for driveability assessment of challenging scenarios.
\end{itemize}

The rest of the paper is organized as follows: Section \ref{sec:factors} presents the factors contributing to driveability and their related studies and challenges. Section \ref{sec:metrics} introduces existing metrics used for driveability assessment and discusses their limitations. Section \ref{sec:datasets} carries out the study of existing public driving datasets. Section \ref{sec:learninggap} and Section \ref{sec:discussions} propose approaches that enable learning when data is scarce. Finally, Section \ref{sec:conclusions} concludes the paper.

\section{Driveability Factors}
\label{sec:factors}
\begin{figure*}[htbp]
\setlength{\abovecaptionskip}{0cm} 
\centering
\includegraphics[trim = 0mm 0mm 0mm 0mm, clip=true,width=18cm]{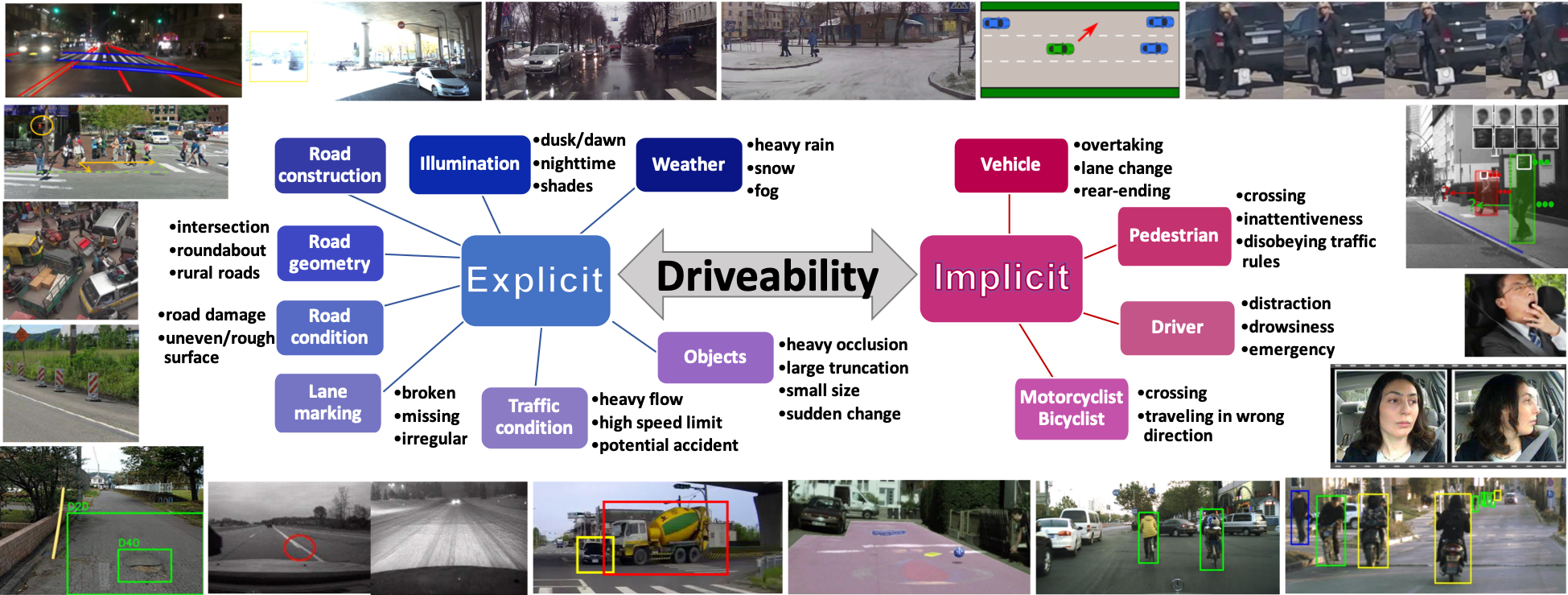} 
\caption{Illustration of factors contributing to driveability assessment and their associated hazardous scenarios. Exemplary figures are adapted from \cite{huang2018apolloscape, seo2015recognition, jain2015car, hillel2014recent, kotseruba2016joint, scheel2018situation, yu2018bdd100k, maeda2018road, pinggera2016lost, suzuki2018anticipating, li2016new, Kang2013VariousAF, okumura2016challenges, japuria2017casnsc}.}
\label{fig:factors}
\end{figure*}

The driveability of a scene is greatly affected by environmental conditions such as weather, traffic flow, road condition, obstacles, and so on, which are explicit factors that can be directly perceived from the environment. However, environmental factors alone are not sufficient for driveability assessment. At times, potential risks can only be identified if the intent and interactions of road users are understood \cite{chowdhury2014user, takahashi2016explaining}, which is implicit information that needs to be inferred from observation. Therefore, in this section, we present both explicit and implicit factors that contribute to driveability and summarize the most relevant elements associated with each factor, which is illustrated in Fig. \ref{fig:factors}. Note that Fig. \ref{fig:factors} only includes elements that are more likely to lower the driveability, which are generally more difficult to handle compared to good or controlled driving conditions assumed in most autonomous driving studies (e.g., \textit{heavy rain}, \textit{snow} and \textit{fog} are included in the ``weather" factor as opposed to \textit{sunny}). Furthermore, these elements are less studied and not well understood by the research community, and therefore require more investigation.

Handling each factor presents its own research challenge and existing approaches are still far from delivering robust solutions that incorporate all factors. To maintain focus over a large research landscape, we will point to surveys on generic studies of these factors and only detail representative works that focus on scenarios with high complexity, dynamics and uncertainty.

\subsection{Explicit Factors}

\subsubsection{Weather/Visibility} Extreme weather such as \textit{fog}, \textit{heavy rain} and \textit{snow} can significantly impair road visibility and sensor performance. Deep neural network (DNN) models are known to behave erroneously under adverse weather conditions \cite{zhang2018deeproad}. There are some studies \cite{gern2002vision, wang2017all} that propose methods that are robust to all weather conditions. However, it is not well understood how various weather conditions affect the reliability of autonomous cars.

\subsubsection{Illumination} Variations in illumination caused by \textit{time of day (dusk/dawn/night)}, \textit{landscape (shades)} and \textit{light sources} pose different challenges for environment perception. \cite{alvarez2008illuminant, corke2013dealing, maddern2014illumination} propose various illumination invariant transforms to improve the robustness of visual perception of autonomous cars under different directions and intensities of the light sources. Nighttime is usually specially handled, where most studies focus on nighttime vehicle detection \cite{chen2006nighttime, kuang2016combining} and rely on vehicle lights as detectable features. However, nighttime pedestrian detection is much harder and more often additional thermal or far infrared data is needed for accurate detection \cite{wagner2016multispectral, gonzalez2016pedestrian}.

\subsubsection{Road Geometry} \textit{Intersections} and \textit{roundabouts} are more difficult to drive through compared to straight highway roads. As pointed out in \cite{7823100}, a significant number of accidents involve intersections. Consequently, intersection is the most studied road geometry, where \cite{shirazi2017looking} makes an overview of recent studies on road users behaviors at intersections. Another notable work is \cite{geiger20143d}, which uses different combinations of visual cues such as scene flow, semantic labels, vanishing points, etc for intersection scene understanding including object detection, driveable area estimation, street orientation and lane detection. Compared to intersections, roundabouts are even more challenging given the stringent time constraints on yielding and merging maneuvers. \cite{okumura2016challenges} reviews related studies and proposes an action planning method to enable an autonomous car to merge into a roundabout. While there are many studies focusing on intersections and roundabouts, almost all consider urban scenarios. In contrast, for rural and unmapped areas, the road geometry and its impact on autonomous driving are not well understood. 

\subsubsection{Road Condition} Potentially hazardous road conditions include \textit{road damage}, \textit{uneven surfaces}, \textit {rough surfaces}, etc. \cite{maeda2018road} introduces a large-scale road damage dataset and implements a convolutional neural network (CNN)-based method to detect eight categories of road damage. Terrain roughness is also important for vehicle control because rough terrain induces shock especially at high speed. Survey \cite{papadakis2013terrain} reviews methods that estimate terrain traversability for unmanned ground vehicles which could potentially be extended to autonomous cars. 

\subsubsection{Road Construction} Road construction can affect multiple road appearances including road geometry and traffic signs as well as driving condition. Usually, there are workers present around a construction site that also need particular attention. Therefore, the boundary of road construction needs to be accurately identified and potential hazards there need to be specifically handled. To this end, \cite{seo2015recognition} focuses on identifying work zone boundary and driving condition change on highways.

\subsubsection{Lane Marking} Lane markings provide important reference for driving, where \textit{broken or missing lane markings} and \textit{irregular lane shapes} cause difficulty in performing lane following or lane change. While lane or road detection is an active research area with a good survey \cite{hillel2014recent} reviewing recent progress, very few studies focus on robustness of these methods. A CNN-based method is proposed in \cite{vokhidov2016recognition} to recognize damaged arrow-road markings which is robust to perspective distortion and partial occlusions, whereas \cite{bao2018unpaved} and \cite{huang2018robustlane} focus on unpaved roads with no markings and stochastic lane shapes, respectively.

\subsubsection{Traffic Condition} In general, highway and urban scenarios require specified learning models due to their different traffic characteristics. Particular attention needs to be paid to situations with extremely \textit{high speed limit}, \textit{heavy flows} and \textit{potential accidents}. However, it is not well understood whether/how the autonomous car's behavior is affected by traffic conditions or when an accident is likely to happen. To this end, \cite{suzuki2018anticipating} presents a near-miss accident dataset and uses a quasi-recurrent neural network (RNN) to predict accidents. 

\subsubsection{Static and Dynamic Objects} Object detection and tracking is the most studied topic with surveys \cite{janai2017computer, sivaraman2013looking, mukhtar2015vehicle, geronimo2010survey, li2016new} reviewing methods and benchmarks on detection of static obstacles, vehicles, pedestrians and cyclists. However, existing methods still demonstrate high error when encountering unseen or hard-to-identify objects especially those of \textit{small size}, \textit{heavy occlusion} and \textit{large truncation} \cite{ohn2017all}. Towards detection of uncommon and unexpected obstacles, \cite{pinggera2016lost} constructs a dataset with small obstacles such as lost cargos and proposes a Bayesian fusion framework that combines CNN with a stereo-system for semantic label prediction. Another work \cite{zhou2017fast} focuses on obstacles that have thin structures such as cables and tree branches, and proposes an edge-based visual odometry technique for detection. While these studies cover several types of uncommon obstacles, there are many more that need investigation such as animals, flying hindrances, sudden obstruction, etc. Furthermore, many of these obstacles (like fallen trees) can cause sudden and unexpected changes to the a-priori map required for vehicle localization. These changes can potentially impair localization accuracy and consequently hinder effective planning of autonomous cars.

\subsection{Implicit Factors}
Implicit factors consist of behaviors and intent of road users interacting with the autonomous car. An interesting survey \cite{ohn2016looking} exhaustively reviewed studies on understanding, modeling, and predicting human agents in three domains, namely, inside the vehicle cabin, around the vehicle, and inside surrounding vehicles. \cite{parkin2016understanding} also provides a literature review on interaction between autonomous cars and other road users. We will not repeat the findings presented in these surveys, but only highlight risky behaviors of vehicles, pedestrians, drivers, motorcyclists and bicyclists, respectively.
\subsubsection{Vehicle Behaviors} 
Potentially hazardous vehicle behaviors include \textit{overtaking}, \textit{lane change}, \textit{rear-ending}, \textit{speed-driving} and \textit{failure to obey traffic laws}, where the first three are studied the most with related works summarized in \cite{sivaraman2013looking, bevly2016lane, xue2018survey}. Vehicle behavior prediction can be hard at high-speed or cluttered urban scenarios. Under these considerations, \cite{dixit2018trajectory} reviews studies on trajectory planning and tracking for high-speed autonomous overtaking, whereas \cite{geng2017scenario, schulz2018interaction} propose scenario-adaptive and intention-aware vehicle trajectory estimation approaches for challenging urban scenarios such as unsignalized urban intersections. Note that vehicle behaviors are tightly connected to the driver's condition. Hazardous vehicle behaviors can also occur when their respective drivers are inexperienced, intoxicated, or limited in physical ability to pay careful attention to everyone on the road.
 
\subsubsection{Pedestrian Behaviors} Pedestrians are the most vulnerable road users. Most accidents happen when a pedestrian is \textit{crossing}, and many seem to result from pedestrians' \textit{inattentiveness} or \textit{failure of compliance with law}. \cite{rasouli2018autonomous} gives a comprehensive overview of factors, methods and challenges on pedestrian behavior studies with both the absence and presence of autonomous vehicles. It is argued that the pedestrians' behavior and their perceived risk of autonomous vehicles may vary depending on numerous factors including demographics as age and gender, dynamic factors such as vehicle speed and distance, as well as social norms and culture. It remains an open question how these factors are interrelated and influential in understanding pedestrians' intent.

\subsubsection{Driver Behaviors}
For partial or high automated vehicles, driver's availability is still required at times when the autonomous car is incapable of making reliable decisions and requests to handover control. Driver behavior analysis is the most mature field with a myriad of studies on driver's activity, intent, alertness, skill and style, and so on \cite{ohn2016looking}. Among these aspects, driver \textit{distraction} and \textit{drowsiness} are two main reasons for traffic accidents. Surveys \cite{Dong2009DriverIM, Kaplan2015DriverBA, Kang2013VariousAF} review methods for driver distraction-and drowsiness-detection using both visual features such as facial expression and eye movement, and non-visual features such as physiological signals and car dynamics, where they show that hybrid measures generate fewer false alarms and higher recognition rate. More investigation is needed on handling emergency situations such as sudden driver impairment under medical conditions.

\subsubsection{Motorcyclist/Bicyclist Behaviors} Compared to other groups of road users, the models and methods for bicyclists/motorcyclist behavior analysis are far more limited due to the lack of datasets. Report \cite{kuehn2015cyclist} shows that the majority of accidents happen when a bicycle appears in front of a vehicle and the driver failed to brake in time due to either obstructed view or the bicyclist traveling in the wrong direction. Substantial efforts are needed for trajectory estimation and intent prediction for motorcyclists and bicyclists.

\subsection{Interrelationship of Driveability Factors}
\label{sec:challenges}
In most cases, the driveability factors are interrelated where the change in one factor could affect the others significantly. For example, how road users behave greatly depend on explicit factors such as weather, road condition, road geometry, and traffic condition. While there are many studies on behavior analysis under different driving contexts, those contexts are generally limited to road geometry such as intersection and traffic condition such as highway traffic versus urban traffic \cite{sivaraman2013looking,xue2018survey}. The interrelationship among other factors are under-exploited.  

A more challenging issue arises from the interaction of different groups of road users. The biggest limitation of existing research on behavior analysis is that most studies only focus on one type of road user (vehicles or pedestrians only). Only a few works investigate the \textit{joint attention} of multiple road users including autonomous cars appearing in the same scene \cite{kotseruba2016joint, rasouli2018joint}. Joint attention in autonomous driving is a complicated problem which involves issues from biological, social and algorithmic perspectives and requires methods for multiple tasks including object detection and tracking, pose estimation and intent prediction \cite{rasouli2018joint}. 

As pointed out in \cite{7823109}, autonomous cars should ideally behave in a way that is comprehensible to humans. They should communicate effectively with pedestrians and cyclists, and react safely to any unpredicted human behaviors. Beyond interaction with pedestrians, it is also important to plan for autonomous cars that account for effects on human drivers because they are expected to share the road for the coming decades. In fact, the cause of many accidents involving autonomous cars could be attributed to human drivers expecting autonomous cars to behave differently \cite{dixit2016autonomous}. In this direction, one pioneering work \cite{sadigh2016planning} uses an inverse reinforcement learning approach to optimize the planning for autonomous vehicles that takes into account the effects on human drivers, which makes the autonomous vehicle more efficient and communicative. However, this study was only carried out on simple simulated scenarios, and more investigation is needed about its applicability in real-world situations. 


\section{Driveability Metrics}
\label{sec:metrics}
Currently, there is no unified performance metric for assessing the driveability of a scene, because driveability assessment involves many tasks in perception and behavior analysis, and the performance metric for each task is tightly related to the underlying model used. In this section, we introduce the most relevant metrics established in existing research for scene driveability evaluation \cite{hecker2018failure} and risk assessment \cite{lefevre2014survey}, and present the design methodology underlying these metrics, with the purpose of encouraging the proposal of novel metrics for driveability assessment. While risk assessment metrics are well established and accepted in studies on ADAS, a metric for scene driveability has only been proposed recently and it is aimed at end-to-end driving policy learning. For risk assessment, we follow the discussions made in \cite{lefevre2014survey} and categorize the metrics according to collision-based risks and behavior-based risks, and we refer interested readers to \cite{lefevre2014survey} for details on the works that utilize these metrics.

\subsection{Scene Driveability}
In \cite{hecker2018failure}, scene driveability is defined by how easy a scene is for an autonomous car to navigate and a scene driveability score is used to measure how likely the car will fail. An end-to-end approach is used to calculate this score. Specifically, an end-to-end driving policy learning model using CNN and long-short-term-memory (LSTM) is first trained to predict driving maneuvers including velocity and steering angle. Then the scene driveability score is calculated based on the discrepancies between the predictions made by the trained driving model and the ground-truth maneuvers. If the score is lower than some manually chosen threshold, the scene is considered ``Hazardous". Otherwise, the scene is considered ``Safe". Once all the scenes are labeled safe or hazardous, another CNN model is trained to predict whether a new scene is safe or hazardous. A similar approach is used in \cite{scheel2018situation}, which uses a bi-directional RNN to classify scenes as safe or unsafe for performing a lane change. However, the limitation of such metrics is that they are defined purely from the model prediction outcome, which is highly model dependent. Furthermore, the end-to-end approach barely provides any insight into what makes a scene hazardous. 

\subsection{Collision-Based Risk}
There are two basic metrics for collision risk computation, namely, binary risk indicator and probabilistic risk indicator. The former only predicts whether or not a collision will happen in the near future, while the latter represents risk score by a probability calculated based on current states, event, choice of hypothesis, future states and damages \cite{eggert2014predictive, 6025208}. Conventional methods that calculate collision-based risk first predict the potential future trajectories for moving entities and then detect collisions between each pair of trajectories \cite{lefevre2014survey}. Newer approaches use deep predictive models to predict whether a collision will occur in the future directly from videos and other sensor data \cite{strickland2017deep}. 

Another widely used indicator is Time-To-X (TTX), where X refers to a relevant event in the course towards collision. The most standard TTX indicator is Time-To-Collision (TTC), which measures the time remaining before the collision occurs and provides clues on whether the car should send a warning to the driver or directly perform an action. Recent study \cite{wachenfeld2016worst} also proposed a worst-time-to-collision (WTTC) metric with the purpose of selecting most critical objects and situations out of a typical test drive to reduce the amount of data saved. An object or situation is considered less critical if its associated WTTC is large. However, the calculation of TTC in most studies relies on simple assumptions of vehicle status and trajectories, which may be difficult to adapt to real-world driving scenarios.

\subsection{Behavior-Based Risk}
Behavior-based risk estimation is usually cast as a binary classification problem, where ``nominal behaviors" are learned from data and then ``dangerous behaviors" are detected. Nominal behaviors are defined based on acceptable speeds, traffic rules, location semantics, weather conditions, and/or the level of fatigue of the driver. For situations involving more than one vehicle, pairs of maneuvers can be labeled as ``conflicting" or ``not conflicting" \cite{lefevre2014survey}. However, behavior-based risk estimation mostly focuses on driver behaviors, while it barely covers the behaviors of other traffic participants. 

\subsection{Challenges in Metric Design}
There are some limitations of existing metrics. First, in most cases, if not all, the ``metrics'' are not strict mathematical metrics and don't satisfy all metric properties. As such, it is difficult to understand and analyse them across systems and scenarios in a commensurable fashion. Second, typically, even though the risk is measured on a continuous scale, for usability these measurements are thresholded (e.g., \textit{Safe} vs \textit{Hazardous}). This results in information-loss. Moreover, they also render the categorization subjective. Third, while the existing metrics can evaluate some types of risks, none of them provides a high-level explanation covering all driveability factors.

Even though not covered in this paper, it is also worth considering whether non-safety factors such as psychological and emotional factors of both the driver and the passengers should also be accounted for by the driveability metric. Ideally, autonomous driving should enable transportation that is not only safe but also enjoyable. To this end, a recent work \cite{7166422} investigated ride comfort measures in autonomous cars, which are shown to be affected by factors such as motion sickness, naturality, apparent safety, etc. Some metrics for evaluation of driver’s emotional feelings such as road frustration index has also been developed in industrial applications. However, it remains a challenging question how to harmonize the concept of driveability for both autonomous cars and the humans inside the vehicle.

As shown above, the actual scope of driveability metric depends greatly on the driving context and use case, which makes it challenging to design a single comprehensive and interpretable metric for driveability. Therefore, composite metrics should be thought of that include both basic measures that are generalizable across systems and additional measures that can be tailored to the specific requirements of target application. Furthermore, such composite metrics could contain heterogeneous measures. Those can include but are not limited to continuous risk scores, the states and trajectories of all traffic participants \cite{sivaraman2014dynamic}, and even semantic descriptors such as “road construction” or “low lane marking visibility” which are shown to better communicate the intent of the autonomous car to its driver \cite{koo2015did}. There is also dependency on the availability of up-to-date a-priori maps of the roads as well as cultural issues specific to a region. For example, driving in the US is probably easier than in some of the less developed countries where roads are less organized and people tend to disobey traffic rules more often.



\section{Datasets}
\label{sec:datasets}
\begin{table*}[htbp]
\scriptsize
\def\sym#1{\ifmmode^{#1}\else\(^{#1}\)\fi}
\caption{Overview of publicly open datasets for autonomous driving.}
\caption*{\footnotesize Abbreviations used: VI-Video, IM-Image, Li-LiDAR, VD-Vehicle Data, CO-Codes; BB-Bounding Box, SL-Semantic Label, LM-Lane Marking, BL-Behavioral Label, O-Others; UR-Urban, RU-Rural, HI-Highway; WE-Weather, SE-Season, LO-Location, NI-Night, IL-Illumination; s-stereo images; uz-unzipped. Unit: K-Kilo, M-Million.}
\centering
\begin{tabular}{ m{2.4cm}| m{2.4cm}| m{2.6cm}| p{0.01cm} p{0.01cm} p{0.02cm} p{0.1cm} p{0.02cm} p{0.01cm}|p{0.01cm} p{0.01cm} p{0.02cm} p{0.1cm} | p{0.01cm} p{0.01cm} p{0.04cm} |p{0.04cm} p{0.02cm} p{0.02cm} p{0.03cm} | m{1.8cm}}
\toprule
\multirow{2}{*}{Dataset}&\multirow{2}{*}{Provider}&\multirow{2}{*}{Time \& Venue} & \multicolumn{6}{c|}{Data provided} & \multicolumn{4}{c|}{Annotation} & \multicolumn{3}{c|}{Traffic} & \multicolumn{4}{c|}{Diversity}&\multirow{2}{*}{Volume}\\
\cline{4-20}
&            &                                                                                                                                    &VI &IM &Li &GPS&VD&CO        &BB&SL&LM&BL       &UR  &RU&HI       &WE&SE&NI & IL&\\
\hline
Apollo Open Platform($\star$)\cite{apollo}& Baidu Inc & 2018; multiple cities in China                                      &\ck&   &\ck&    &\ck&            &\ck &     &    &           &\ck&\ck&\ck        &\ck&   &    &   &            multiple datasets, volumes vary\\
\hline 
ApolloScape\cite{huang2018apolloscape} & Baidu Inc & 2018; multiple cities in China                                                                  &\ck&   &   &    &    &             &    &\ck&\ck&             &\ck&\ck&\ck       &\ck&   &    &\ck&            $>$140K IM \\
\hline
Belgium Traffic Sign\cite{timofte2014multi} & ETH Z{\"u}rich & 2011; Belgium                                                                          &    &\ck&   &    &    &            &\ck&     &    &             &\ck&    &  	      &     &    &     &    &            $>$9K IM, 50GB      \\
\hline
Berkeley DeepDrive\cite{yu2018bdd100k} & UC Berkeley & 2017; San Francisco Bay Area, New York, US                      &\ck&\ck&   &\ck&    &\ck       &\ck&\ck&\ck&            &\ck&\ck&\ck       &\ck&    &\ck&\ck&            100K IM, 1.8TB \\
\hline 
Bosch Small Traffic Lights\cite{behrendt2017deep} & Bosch North America Research & 2017; San Francisco Bay Area, US   &\ck&   &   &    &    &              &\ck&     &    &           &\ck&\ck&            &\ck&    &      &\ck&          $>$13K IM\\
\hline
Brain4Cars\cite{jain2015car} &Cornell Univ. &2016; two states in US							                     &\ck&    &   &\ck&\ck&\ck       &     &     &    &\ck       &\ck&    &\ck        &     &    &     &\ck&            700 VI\\
\hline
Caltech Pedestrian\cite{dollar2009pedestrian} & California Inst. of Tech. & 2009; Los Angeles, US                                           &\ck&   &   &    &    &\ck          &\ck&     &    &           &\ck&    &             &     &    &     &    &             250K IM, 11GB\\
\hline
CamVid\cite{brostow2008segmentation} & Univ. of Cambridge & 2009; Cambridge, UK 							             &\ck&\ck&   &   &    &\ck         &     &\ck&     &           &\ck&    &             &     &    &     &    &             700 IM, 8GB\\
\hline
CCSAD\cite{guzman2015towards} & Centro de Investigacin en Matemticas & 2014; Guanajuato, Mexico				     &\ck&   &   &\ck&    & 	          &     &     &    &            &\ck&    &              &     &    &\ck&\ck&           $>$96K IM(s), 500GB\\
\hline
CityScapes\cite{cordts2016cityscapes}&Daimler AG,MPI-IS,TU Darmstadt & 2016; 50 cities in Germany, Switzerland \& France    &\ck&\ck&   &\ck&\ck&\ck       &     &\ck&     &    &\ck&    &       &     &\ck&     & &         25K IM(s), 63GB\\
\hline
CMU\cite{badino2012real} & Carnegie Mellon Univ. & 2011; Pittsburgh, US 							              &\ck&   &\ck&\ck&    &            &     &     &    &            &\ck&    &               &\ck&\ck   &    &\ck&          16 VI, 275GB\\
\hline
Comma.ai\cite{santana2016learning} & comma.ai & 2016; San Francisco, US									     &\ck&   &    &\ck&\ck&\ck       &     &     &    &            &    &    &\ck           &     &    &\ck&   &        80GB\\
\hline
CULane\cite{pan2018SCNN} & Chinese Univ. of Hong Kong & 2018; Beijing, China                                                        &     &\ck&   &    &     &\ck       &    &     &\ck& 	    &\ck&\ck&\ck          &     &    &\ck&\ck&            133K IM\\
\hline
Daimler Pedestrian($\star$)\cite{enzweiler2008monocular} & Daimler AG R\&D, Univ. of Amsterdam &2006-2016; Beijing, China, others unknown    &\ck&\ck&   &   &    &		  &\ck&\ck&    &	             &\ck&    & 		  &     &    &     &    & 		   8 datasets, 2.5MB - 45GB each\\
\hline
DAVIS\cite{binas2017ddd17} & Univ. of Z{\"u}rich, ETH Z{\"u}rich& 2017; Switzerland, Germany                                &\ck&   &    &\ck&\ck&\ck        &     &     &    &            &\ck&\ck&\ck           &\ck&    &\ck&\ck&               41 VI, 450GB\\
\hline
DBNet\cite{chen2018lidar} & Shanghai Jiao Tong Univ., Xiamen Univ. & 2018; several cities in China                              &\ck&   &\ck&   &\ck& 		  &     &     &    &	      &\ck&\ck&\ck          &     &    &     &    &             $>$10K IM, $>$1TB(uz)\\
\hline
DIPLECS($\star$)\cite{pugeault2015much} & Univ. of Surrey & 2015; Surrey, UK, Stockholm, Sweden						       &\ck&   &    &   &\ck&            &     &     &    &\ck	      &\ck&\ck&\ck		    &     &    &     &    &             4.3GB \& 1.1GB\\
\hline
Dr(eye)ve\cite{alletto2016dr} & Univ. of Modena and Reggio Emilia & 2016; Modena, Italy										       &\ck&   &    &   &\ck&	  &     &     &    &\ck	       &\ck&\ck&\ck          &\ck&    &\ck&   &  		555K\\
\hline
EISATS($\star$)\cite{EISATS} & Univ. of Auckland & 2010; multiple locations in Germany and New Zealand   &\ck&   &\ck&   &   &    &     &     &    &\ck      &\ck&\ck&\ck   &\ck&    &\ck&\ck&   multiple datasets, volumes vary\\
\hline
Elektra($\star$)\cite{Elektra} & Autonomous Univ. of Barcelona, Polytechnic Univ. of Catalonia&2016; Barcelona, Spain  &\multicolumn{6}{c|}{VI \& Infrared}  &\ck&     &    &	       &\ck&    &               &     &    &\ck&   &                 9 datasets, 0.5GB-12GB each\\
\hline
ETH Pedestrian\cite{eth_biwi_00534} & ETH Z{\"u}rich & 2009; Z{\"u}rich, Switzerland 						&\ck&   &   &    &    &           &\ck&     &    &              &\ck&    &   	 &     &    &     &    &                   $>$4.8K IM(s), 660MB\\
\hline
Ford\cite{pandey2011ford} & Univ. of Michigan & 2009; Michigan, US										&\ck&  &\ck&\ck&   &\ck		   &     &     &    &             &\ck&    &             &     &    &     &    &        80GB \& 120GB\\
\hline
German Traffic Sign\cite{Houben-IJCNN-2013} & Ruhr Univ. Bochum & 2012; Germany							&     &\ck&   &    &    &                   &\ck&     &    &             &    &    &              &     &    &     &    &       5K IM, 1.6GB\\
\hline
HCI Challenging Stereo\cite{meister2012outdoor} & HCI(Heidelberg), Bosch Corporation Research& 2012; Hildesheim, Germany  &\ck&  &   &    &    &        &     &     &    &             &\ck&    &              &\ck&   &\ck&\ck&       11 VI(s)\\
\hline
HD1K\cite{HD1K} & HCI(Heidelberg), Robert Bosch GmbH& 2018; Heidelberg, Germany                             &\ck&   &    &    &     &  &\multicolumn{4}{c|}{optical flow}  &\ck&    &            &\ck&   &\ck&\ck&        $>$1K\\
\hline
Highway Workzones\cite{seo2015recognition} & Carnegie Mellon Univ. & 2015; US                                     &\ck&   &    &    &     &           &\ck&     &   &              &&    &\ck                        &\ck&\ck   &&\ck&        6 VI, 1.2GB\\
\hline
JAAD\cite{kotseruba2016joint} & York University& 2016; mostly in Ukraine and Canada                        &\ck&   &    &    &     &                      &\ck&     &    &\ck         &\ck&    &             &\ck&    &\ck&    &        347 VI, 170GB\\
\hline
KAIST Multi-Spectral\cite{choi2018kaist}&KAIST &2015; South Korea                                                                       &\multicolumn{6}{c|}{VI, Li, GPS, CO, Thermal}  &\ck&     &    &   &\ck&    &    &     &    &\ck&\ck&       10 VI\\
\hline
KAIST Urban\cite{jeong2018dataset}&KAIST&2018; multiple cities in South Korea						       &\ck &&\ck&\ck&    &\ck                  &     &     &    &             &\ck&    &                &     &    &     &    &       19 VI, 1GB-22GB each\\
\hline
KITTI($\star$)\cite{Geiger2013IJRR} & Karlsruhe Inst of Technology, Toyota Technological Inst&2011; Karlsruhe, Germany  &\ck&\ck&\ck&\ck&    &\ck         &\ck&\ck&\ck&		&\ck&\ck&\ck          &     &    &     &\ck&         multiple datasets, volumes vary\\
\hline
LISA Traffic Sign\cite{mogelmose2012vision} & Univ. of California, San Diego & 2012; US							&     &\ck&   &    &    &\ck                  &\ck&     &    &           &    &    &                &     &    &     &    &          6.6K IM, 8GB\\
\hline
LostAndFound\cite{pinggera2016lost} & Daimler AG&2016, Germany										&     &\ck&   &    &\ck&                      &     &\ck&     &           &\ck&    &                &     &    &     &    &         2K IM(s), 40GB\\
\hline
M{\'a}laga\cite{blanco2014malaga} &Univ. of M{\'a}laga&2014; M{\'a}laga, Spain								&\ck&    &\ck&\ck&    &\ck                &     &     &    &           &\ck&    &                &     &    &     &    &          15 VI(s), 70GB\\
\hline
Mapillary Vistas\cite{neuhold2017mapillary} & Mapillary AB & 2017; around the globe								&     &\ck&   &    &    &		       &     &\ck&     &          &\ck&\ck&                &\ck&\ck &\ck&\ck&           25K IM\\
\hline
NEXET\cite{Nexar} &Nexar &2017; around the globe											&     &\ck&   &    &    &\ck                  &\ck&     &    &          &\ck&\ck&\ck             &\ck&\ck &\ck&\ck&          55K IM, 10GB\\
\hline
nuScenes\cite{nuscenes} & nuTonomy Inc, Aptiv & 2018; Boston, US \& Singapore                                                              &\ck &\ck&\ck&\ck &    &\ck          &\ck&     &    &\ck      &\ck&    &          &\ck&  &    &\ck&               1K VI, 40K IM\\
\hline
Oxford RobotCar\cite{maddern20171}& Oxford Univ. &2015; Central Oxford, UK							&\ck&  &\ck&\ck&\ck&\ck                   &     &     &    &           &\ck&    &                &\ck&\ck &    &\ck&           130 VI(s), 23TB\\
\hline
Road Damage\cite{maeda2018road}& Univ. of Tokyo & 2018; multiple cities in Japan							&     &\ck&   &    &    &\ck                  &\ck&     &    &            &\ck&    & 		&\ck&   &    &    &              9K IM\\
\hline
Stanford Track\cite{teichman2011towards}&Stanford Univ.&2010; Stanford Univ., US						                &     &  &\ck&\ck&     &\ck                 &\multicolumn{4}{c|}{stixel}  &\ck&    &         &     &    &     &    &           14K tracks, 5.7GB\\
\hline
Stixel\cite{pfeiffer2013exploiting} &Daimler AG&2013, Germany												&\ck&   &   &     &\ck&                &\multicolumn{4}{c|}{stixel}  &    &    &\ck           &\ck&   &    &    &                2.5K IM, 3GB\\
\hline
TME Motorway\cite{TMEMotorwayDataset} & Czech Technical Univ. &2011; Northern Italy							&\ck&   &   &     &    &\ck			&\ck&     &    &             &    &    &\ck           &     &    &     &\ck&             28 VI\\
\hline
TUD-Brussels Pedestrian\cite{wojek2009multi} &Max Planck Inst for Informatics &2009; Belgium				&\ck&   &   &     &    &	                 &\ck&     &    &            &\ck&    &                &     &    &     &    &             1.6K IM\\
\hline
TuSimple($\star$)\cite{TuSimple} & TuSimple &2017; venue unknown							        &\ck&\ck&   &     &    &\ck                    &\ck&     &\ck&           &    &    &\ck           &\ck&   &    &    &                7K IM \& 5K IM\\
\hline
UAH\cite{romera2016need} &  University of Alcal{\'a}&2016; Madrid, Spain                                                        &\ck& &   &\ck&\ck&\ck                              & &     &    & \ck              &    &    &\ck             &     &    &     &    &                35 VI   \\
\hline
Udacity($\star$)\cite{Udacity} & Udacity & 2016; Mountain View, US								&\ck&   &\ck&\ck&\ck&\ck                    &\ck&     &    &            &\ck&    &               &    &   &    &\ck& 		300GB(uz)\\
\bottomrule
\end{tabular}
\label{table:datasetall}
\end{table*}

The development of robust autonomous driving models depends on having access to large-scale training datasets, especially as more learning-based approaches are incorporated. Over the past decade, tens of datasets for autonomous driving have been collected and made public by multiple institutes around the world. These datasets are a valuable resource for the research community to develop benchmarks and consolidate research efforts. However, as autonomous driving encompasses numerous tasks in perception, localization and behavior analysis and the datasets are greatly varied in application focus, it is not trivial to determine which dataset to use for which task. Therefore, in this section, we provide an up-to-date exhaustive list of 45 publicly available datasets for autonomous driving and categorize them according to the tasks that they are suitable for. Particularly, since we have identified the factors that contribute to driveability and their associated challenges in Section \ref{sec:factors}, we will highlight the datasets that can be used to address these challenges. We also publish this dataset archive at \textit{https://sites.google.com/view/driveability-survey-datasets}, which allows interactive exploration of datasets and will continue to be maintained after publication.

Prior to this work, a comprehensive survey of publicly available datasets was published in \cite{yin2017use} and includes 27 publicly available datasets collected on public roads before late 2016. However, some of the largest and most diversified datasets have been released in the past two years, which are not included in \cite{yin2017use}. We enrich the list in \cite{yin2017use} by adding 23 more datasets most of which were released after 2016. We also exclude 5 datasets in \cite{yin2017use} that have broken web links, impose a charge, or are integrated into newer datasets. We would like to mention that besides our selected datasets, there are other reported efforts on dataset acquisition such as the dataset ``TorontoCity" collected by researchers at University of Toronto \cite{wang2017torontocity}. Unfortunately, this dataset no longer supports open access and is therefore not included in this paper. 

Different from \cite{yin2017use} which compares the datasets' metadata such as venue, volume, traffic condition, sensor setup and file type, we compare the datasets from an application perspective in terms of which tasks each dataset is suitable for, diversity and modality, level of annotations, and whether training/testing data and benchmarks are provided. Furthermore, we will summarize the trends emerging in these recently published datasets and propose directions for future dataset collection.

\subsection{Overview of Datasets}
We follow the same inclusion criteria to select relevant datasets as those used in \cite{yin2017use}, i.e., the dataset must be collected by on-board sensors of a vehicle running on public roads, contain camera or LiDAR data, and allow free open access. By an extensive search and snowballing in dataset websites, publications and competitions on autonomous driving, 45 datasets have been included in this survey and their metadata are presented in Table \ref{table:datasetall}. The symbol ($\star$) denotes that the dataset contains multiple subsets that are collected with different sensors and for different purposes. In the following, we elaborate on the aspects considered when presenting these datasets. While license information is not included in Table \ref{table:datasetall}, we suggest the readers check and observe the license when using these datasets.

\textit{Time \& Venue} denotes when the dataset was published and where the data was collected. In terms of time, half of the datasets were released since the year of 2016, which shows an increasing interest and effort on public dataset collection to boost the progress in autonomous driving research. In terms of venue, while most datasets published before 2016 were collected in Europe and the United States, the collection of many of the more recent datasets took place in Asian countries (such as ApolloScape, DBNet, KAIST, Road Damage Dataset, etc), and even around the globe (Mapillary Vistas and NEXET) with collaborative efforts from drivers worldwide who uploaded images to the database. Such diversity in locations is a leap forward towards enabling autonomous driving on road networks all over the world.

\textit{Data provided} lists the data modalities and development codes provided by the dataset. A dataset is considered to contain videos if it provides either videos or image sequences that capture temporal information. Otherwise, a dataset is considered to only contain images if standalone images are provided without preceding or tailing video clips. For datasets that contain raw video clips with annotations made on selected but not all video frames, we mark them as containing both video and image data. Additional to videos and images, around one-fourth of the presented datasets provide LiDAR scans, GPS/IMU (inertial measurement unit) data, and/or vehicle status data including steering angle and velocity. There are also two datasets that provide infrared and thermal scans. Data collected from various types of sensors is greatly appreciated in deep multimodal learning, which is believed to enhance the inference performance of deep neural networks \cite{ramachandram2017deep}. Most of the datasets provide codes in Python, Matlab or C++ for data preprocessing and visualization. 

\begin{figure*}[t]
\setlength{\abovecaptionskip}{0cm} 
\centering
\captionsetup[subfigure]{captionskip= 0 cm}
\subfloat[bounding box]
{
\label{fig:bounding box}
\includegraphics[trim = 0mm 0mm 0mm 0mm, clip=true,width=4.2cm, height = 2.2cm]{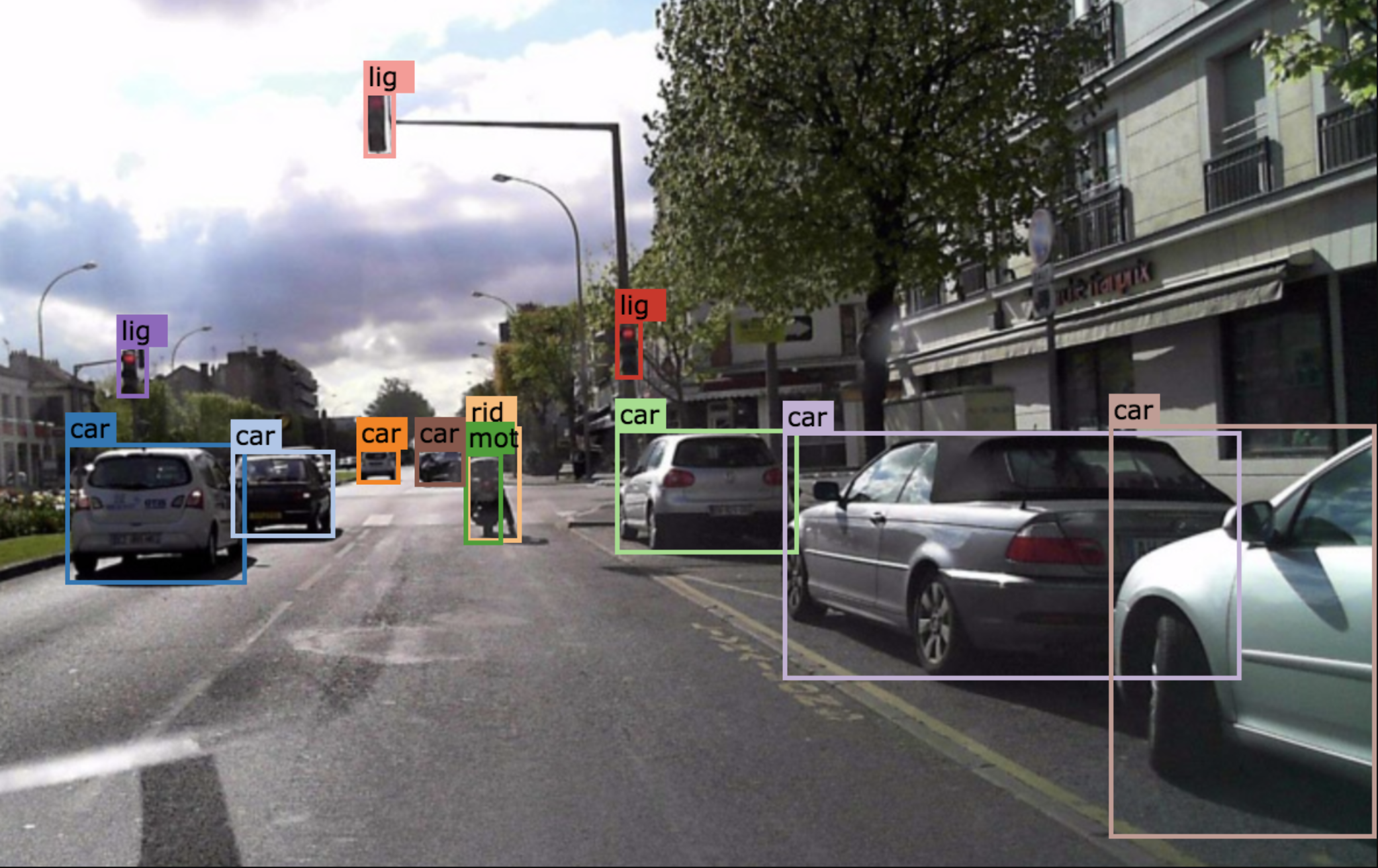} 
}
\hspace{0cm}
\subfloat[semantic label]
{
\label{fig:semantic}
\includegraphics[trim = 0mm 0mm 0mm 0mm, clip=true,width=4.2cm, height = 2.2cm]{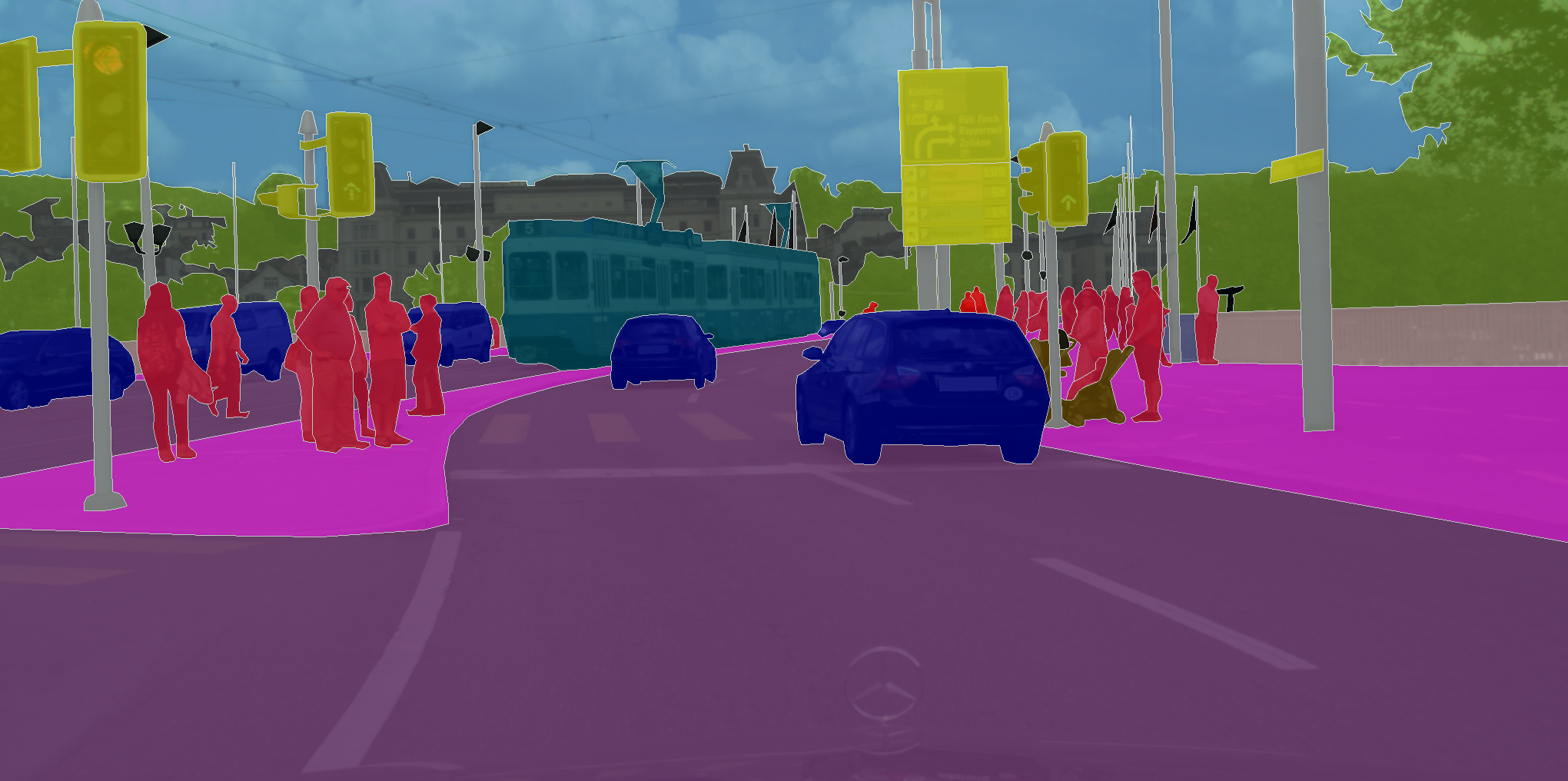} 
}
\hspace{0cm}
\subfloat[lane marking]
{
\label{fig:lane_marking}
\includegraphics[trim = 0mm 0mm 0mm 0mm, clip=true,width=4.2cm, height = 2.2cm]{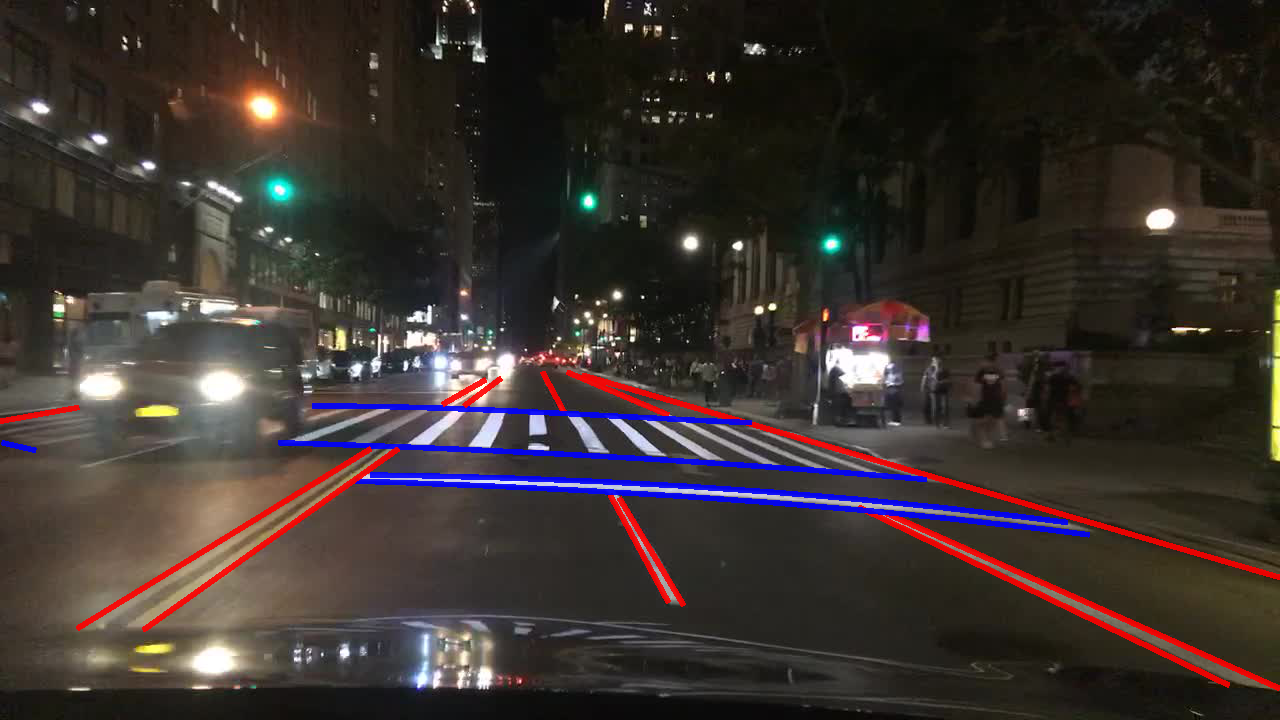} 
}
\hspace{0cm}
\subfloat[driveable area]
{
\label{fig:drivable}
\includegraphics[trim = 0mm 0mm 0mm 0mm, clip=true,width=4.2cm, height = 2.2cm]{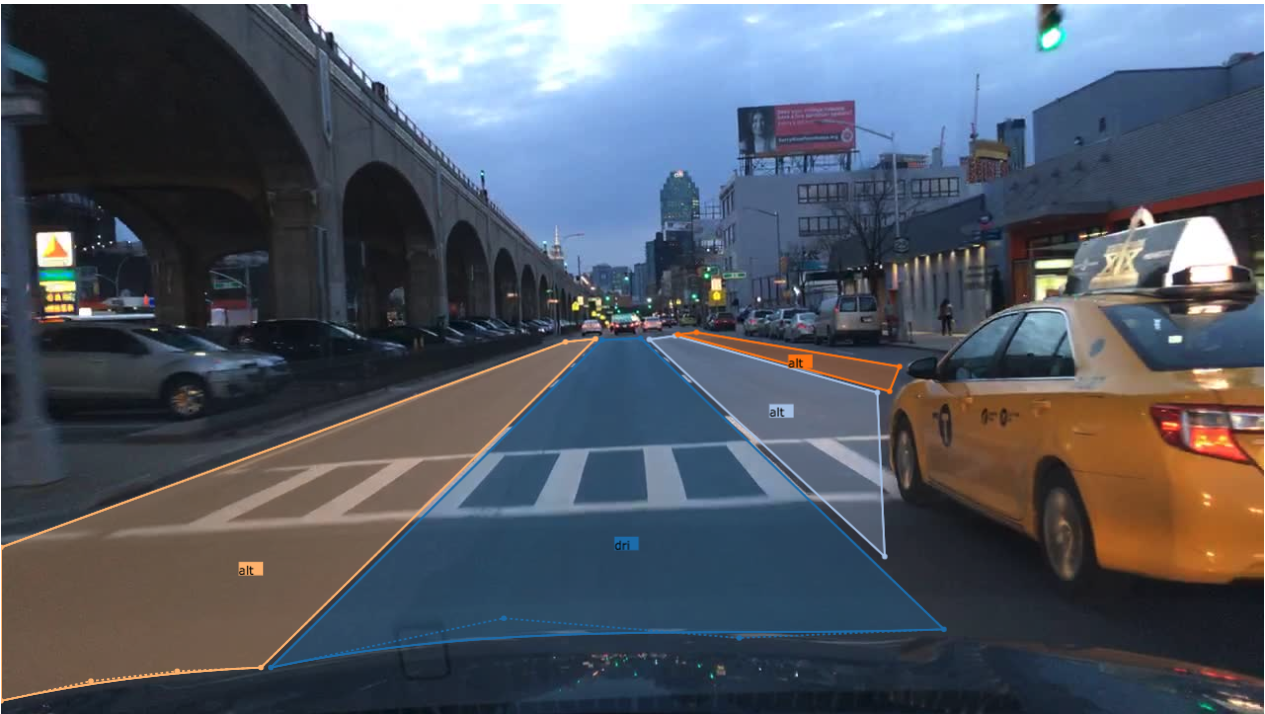} 
}\\
\subfloat[driver's gaze]
{
\label{fig:gaze}
\includegraphics[trim = 0mm 0mm 0mm 0mm, clip=true,width=4.2cm, height = 2.2cm]{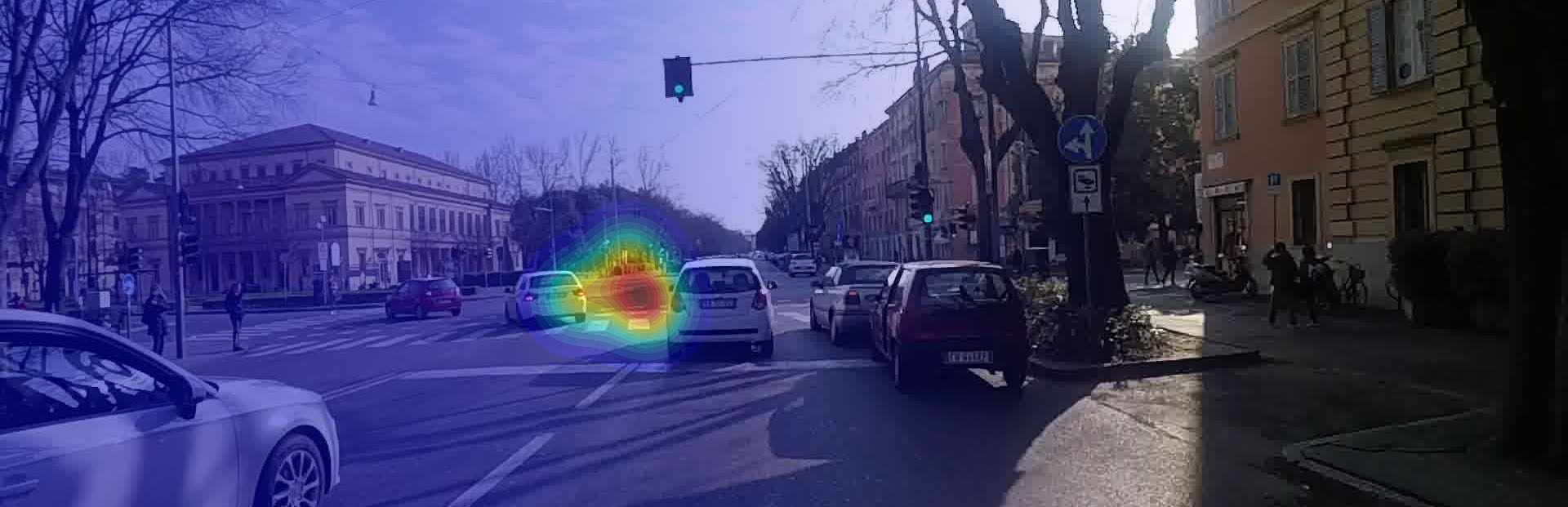} 
}
\hspace{0cm}
\subfloat[stixel label]
{
\label{fig:stixel}
\includegraphics[trim = 0mm 0mm 0mm 0mm, clip=true,width=4.2cm, height = 2.2cm]{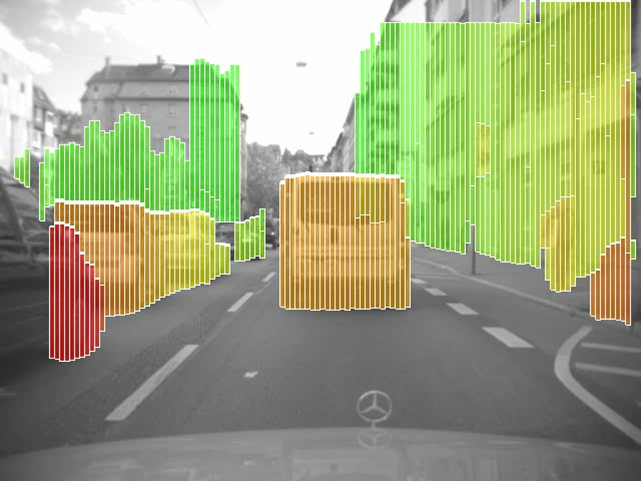} 
}
\hspace{0cm}
\subfloat[depth image]
{
\label{fig:depth}
\includegraphics[trim = 0mm 0mm 0mm 0mm, clip=true,width=4.2cm,height = 2.2cm]{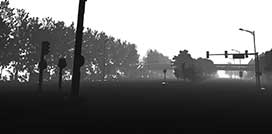} 
}
\hspace{0cm}
\subfloat[optical flow]
{
\label{fig:optical_flow}
\includegraphics[trim = 0mm 0mm 0mm 0mm, clip=true,width=4.2cm, height = 2.2cm]{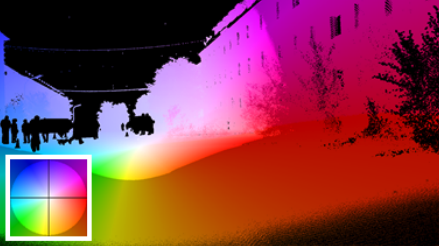} 
}
\caption{Sample images with various annotation types. (a)(c)(d) are from Berkeley DeepDrive, (b) is from CityScapes, (e) is from Dr(eye)ve, (f) is from Stixel, (g) is from ApolloScape, and (h) is from HD1K. All sample images are extracted from the webpages hosting the datasets.}
\label{fig:annotations}
\end{figure*}

\textit{Annotation} presents the type of labels provided by the dataset. The most common annotation types are bounding boxes and semantic labels, where the former is used for object detection and tracking and the latter is used for semantic segmentation. Some datasets also provide lane markings for lane detection. In addition to these graphical labels, a few datasets provide behavioral labels or data for higher-level scene reasoning. Such labels include synchronized videos of both the driver's face and the road (Brain4Cars), driver's gaze (Dr(eye)ve, Elektra) and behaviors of both the drivers and pedestrians present in the same scene (JAAD). Besides these annotations, the optical flow information is provided by the HD1K dataset, and a stixel label that uses multiple cubics to represent an object is provided in the Stanford Track and the Stixel dataset. Figure \ref{fig:annotations} shows one example for each type of annotation. Data labeling is generally considered to be time-consuming and cumbersome, and therefore choosing a dataset with annotations could ease the labeling burden in model training.

\textit{Traffic} records the traffic conditions where the majority of datasets focus on urban traffic. \textit{Diversity} shows the diversity in environmental conditions under which the dataset was collected, which includes \textit{weather, season, night and illuminations}. Two-thirds of the presented datasets provide diversified data in at least one of the above categories. However, only two datasets, namely, Mapillary Vistas and NEXET, demonstrate diversity across all four categories. Seasonal changes are not commonly seen due to the fact that most datasets are collected over shorter durations. 

\textit{Volume} shows the total dataset size in terms of zipped files if not specified otherwise, which provides reference for disk usage when one considers storing and utilizing the dataset. 

\begin{table}[tb]
\caption{Dataset categorization by autonomous driving tasks.}
\caption*{\footnotesize \ck denotes that separate training and test sets are provided; $\bigstar$ denotes that benchmark results are provided.}
\centering
\begin{tabular}{ m{1.3cm}| m{6.7cm}}
\toprule
Task & Datasets\\
\hline
Stereo / 3D vision & CamVid, CCSAD, CMU, EISATS, Elektra, HCI Challenging Stereo, KAIST Urban, KITTI(\ck $\bigstar$), M{\'a}laga, Oxford Robotcar, Stixel\\
\hline
Optical flow & HCI Challenging Stereo, HD1K, KITTI(\ck $\bigstar$)\\
\hline
\multirow{4}{1em}{Object detection} & Multi-class: Apollo Open Platform, Berkeley DeepDrive (\ck $\bigstar$), CamVid, CityScapes(\ck), Elektra, JAAD, KAIST Multi-Spectral, KAIST Urban, KITTI(\ck $\bigstar$), NEXET(\ck), nuScenes, Stanford Track, TME Motorway, Udacity(\ck)\\
&Traffic sign (TS): Belgium TS(\ck), Bosch Small Traffic Lights(\ck $\bigstar$), Highway Workzones, German TS(\ck $\bigstar$), LISA TS\\
& Pedestrian (Ped): Caltech Ped(\ck $\bigstar$), Daimler Ped(\ck $\bigstar$), ETH Ped, TUD-Brussels Ped(\ck)\\
& Obstacle: LostAndFound(\ck $\bigstar$), Road Damage(\ck $\bigstar$)\\
\hline
Object tracking & Apollo Open Platform, Berkeley DeepDrive(\ck), Caltech Ped(\ck), Daimler Ped, Elektra, ETH Ped, JAAD, KAIST Multi-Spectral, KITTI(\ck $\bigstar$), nuScenes, Stanford Track, TME Motorway, TUD-Brussels Ped(\ck), TuSimple(\ck), Udacity(\ck)\\
\hline
Lane/Road detection & ApolloScape(\ck), Berkeley DeepDrive(\ck), CULane(\ck $\bigstar$), KAIST Multi-Spectral($\bigstar$), KITTI(\ck $\bigstar$), TuSimple(\ck $\bigstar$)\\
\hline
Semantic segmentation & ApolloScape(\ck $\bigstar$), Berkeley DeepDrive(\ck), CamVid, CityScapes(\ck $\bigstar$), Daimler Ped(\ck), Elektra, KITTI(\ck $\bigstar$), Mapillary Vistas(\ck $\bigstar$)\\
\hline
Localization / SLAM & CMU, Ford, KAIST Multi-Spectral, KAIST Urban, KITTI(\ck $\bigstar$), M{\'a}laga, Oxford Robotcar, Udacity(\ck $\bigstar$)\\
\hline
End-to-end learning & Apollo Open Platform, Comma.ai, DAVIS, DBNet($\bigstar$), DIPLECS, Udacity(\ck $\bigstar$)\\
\hline
Behavior analysis & Brain4Cars($\bigstar$), DIPLECS, Dr(eye)ve, EISATS, Elektra, JAAD, UAH\\
\bottomrule
\end{tabular}
\label{table:datasettask}
\end{table}

\subsection{Dataset Categorization}
\begin{figure}[tbp]
\setlength{\abovecaptionskip}{0cm} 
\centering
\includegraphics[trim = 0mm 0mm 0mm 0mm, clip=true,width=8.5cm]{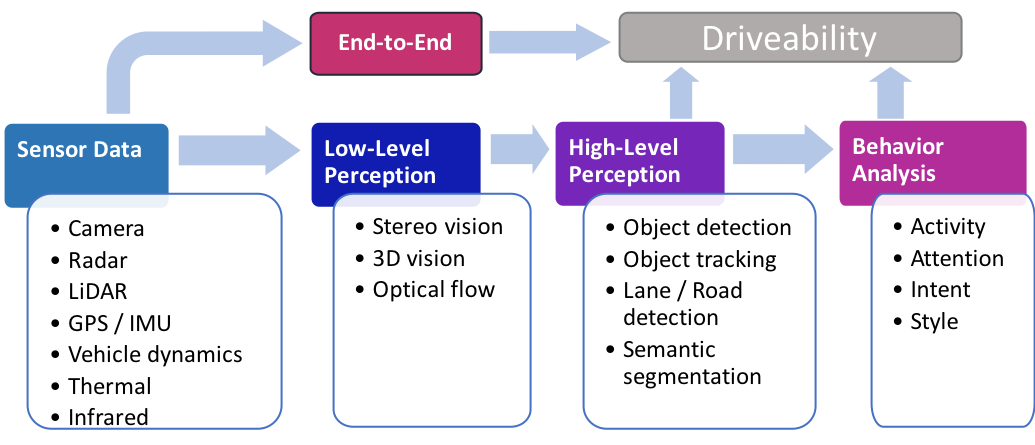} 
\caption{Overview of the learning pipeline for driveability assessment.}
\label{fig:tasks}
\end{figure}

As presented in Table \ref{table:datasetall}, nearly no two datasets are exactly the same in terms of what they offer. The datasets are usually collected and annotated for different purposes, and therefore, they are suitable for different autonomous driving tasks. To provide a practical view of how to use these datasets, we categorize the datasets according to tasks. In Table \ref{table:datasettask}, we list common autonomous driving tasks based on the autonomous driving frameworks released by both research institutes \cite{ulbrich2017towards, lin2018architectural} and industrial companies \cite{Mobileye, ApolloStack}, and the tasks included in KITTI benchmarks \cite{KITTI}. We refer interested readers to \cite{janai2017computer} for detailed descriptions of these tasks. The role of related tasks in the learning pipeline for driveability assessment is illustrated in Fig. \ref{fig:tasks}.  Even though for a specific task, we only include datasets that were originally collected to perform the task, it should be noted that other datasets may also be applicable with additional labeling or data processing. We also show whether a dataset includes separate training and test sets, and whether the providers offer benchmark results online or in publications, which are very useful for comparison of methods developed using the dataset. 

From Table \ref{table:datasettask}, it can be observed that most of the datasets were collected for object detection, which is one of the early research fields for autonomous driving. There are also a good collection of datasets for object tracking and semantic segmentation. However, only few datasets can be used for reasoning about a scene at a higher level using behavior analysis, which is a research topic that has been studied insufficiently. Furthermore, as the end-to-end approach has been shown to have full potential of driving policy learning \cite{bojarski2016end}, a couple of datasets were published recently to encourage research in this direction. 

\subsection{Dataset Highlights for Driveability Assessment}
While classical datasets such as KITTI provide canonical benchmarks for testing and comparison of baseline algorithms, datasets that capture more adverse scenes or provide different levels of annotations will be more interesting for improving the robustness of autonomous cars. The collection of challenging scenes helps to identify the limitations of state-of-the-art approaches and inspire novel and more robust algorithms. Fortunately, towards this goal, the following trends have been observed in recently released datasets, namely, high complexity and diversity, capturing potentially hazardous events, and providing behavioral and contextual data for prediction and inference. In the following, we highlight the datasets according to these three characteristics and take a deeper dive into what is offered in each dataset. 
\subsubsection{High Diversity and Complexity}
To capture realistic driving scenes, the data collected should be as natural and diversified as possible in terms of weather, traffic condition, illumination, etc. The most diverse datasets are Mapillary Vistas and NEXET, which consist of images uploaded from drivers all over the globe covering six continents. NEXET was collected by using the Nexar’s dashcam and almost half of the images were taken at night time \cite{Nexar}. Bounding boxes for five classes of vehicles are provided. Compared to NEXET, Mapillary Vistas is also diverse in terms of image sources as the images were taken from different devices including both cameras and mobile phones and by photographers of varying experiences. However, only images with at least 1920$\times$1080 resolution were selected \cite{neuhold2017mapillary}. Semantic labels for 66 classes of objects are provided by Mapillary Vistas in the research edition, whereas the commercial edition provides labels for 100 classes. 

Another two large datasets for semantic segmentation are ApolloScape and CityScapes. CityScapes has gained significant popularity since its release. It provides 5K stereo images with pixel-level semantic labels and 20K stereo images with instance-level semantic labels \cite{cordts2016cityscapes}. The annotated image is taken from a 1.8-second video clip and the raw video clips are provided. 30 classes of objects are annotated and additional information such as outside temperature and precomputed disparity depth maps are also provided. However, the limitation of CityScapes is that most images were collected in daytime under good to medium weather condition. ApolloScape is by far the largest dataset with semantic labels covering 25 classes. Additionally, it provides 28 lane marking classes and depth images. Compared to CityScapes, ApolloScape captures more scenes with bad weather, reflections on vehicles and extreme lighting conditions \cite{huang2018apolloscape}. The largest dataset that provides the most sensor measurements is nuScenes, which contains 1000 20-second long videos with LiDAR, Radar, camera, IMU and GPS data. It also provides 3D bounding boxes over 25 classes of objects annotated at 2Hz. 

Berkeley DeepDrive provides the most comprehensive annotations with bounding boxes for 10 classes and lane markings for 8 classes on 100K images, and semantic labels for 40 classes on 10K images. The raw videos of 40 seconds are provided where the annotated images are extracted. Additionally, a unique annotation of driveable areas and a tag that shows the weather, time of day and scene context (residential, highway, tunnel, etc) of each image are also provided \cite{yu2018bdd100k}.

While most datasets were collected during a short time span, the CMU and Oxford RobotCar datasets were collected over months traversing the same routes multiple times, which capture long-term changes in the environment. Particularly, Oxford RobotCar was collected on the same route twice a week over a year and provides various types of data including stereo and monocular images, 2D and 3D LiDAR scans, GPS, and inertial sensor data, which is good for research on long-term simultaneous localization and mapping (SLAM) in dynamic urban environments \cite{maddern20171}.

\begin{table}[tb]
\caption{Number of instances in training and validation sets.}
\centering
\begin{tabular}{ m{2.4cm}<{\centering}|m{0.7cm}<{\centering}| m{0.7cm}<{\centering}| m{1.4cm}<{\centering}|m{1.4cm}<{\centering}}
\toprule
Dataset & \multicolumn{2}{c|}{Total ($\times 10^3$) } & \multicolumn{2}{c}{Average per image}\\
\cline{2-5}
      &Person&Vehicle&Person&Vehicle\\
\hline
ApolloScape & 543 & 1989 & 1.1/6.2/16.9 &12.7/24.0/38.1\\
\hline
CityScapes & 24.4 & 41&7.0&11.8\\
\hline
KITTI & 6.1&30.3&0.8&4.1\\
\hline
Berkeley DeepDrive &86 & - &1.2&-\\
\hline
Caltech & 192 & - & 1.5 & - \\
\bottomrule
\end{tabular}
\label{table:datacomplexity}
\end{table}

Scenes recorded in cluttered urban environments with more traffic participants are also favorable to enhance the robustness of learning models to deal with complex scenarios. Table \ref{table:datacomplexity} provides a comparison of the number of instances labeled in some large datasets which make this data available. ApolloScape is divided into three complexity levels with different numbers of instances per image \cite{huang2018apolloscape}. ApolloScape and CityScapes contain images with more people as most of the data was collected in urban areas, whereas KITTI and Berkeley DeepDrive contain many scenes on highways with fewer people.

\subsubsection{Hazardous Events}
As identified in Section \ref{sec:factors}, hazardous factors include road damages, construction, adverse weathers, hard-to-identify obstacles due to size or occlusion, and so on. Several datasets have been collected to address these challenges. In terms of road condition, Road Damage is the only dataset that annotates road surface damages. The images were taken by a vehicle-mounted smartphone and 8 types of road damages were identified according to the Road Maintenance and Repair Guidebook in Japan including liner crack, alligator crack, bump, line blur, etc \cite{maeda2018road}. CCSAD was also collected in locations with hazardous road conditions including irregular speed humps, abundant potholes, and peculiar flows of pedestrians \cite{guzman2015towards}. However, only raw data is provided in CCSAD. For road construction, Highway Workzones annotates signs specifically related to workzones, which is, however, limited to only highway traffic.

Among the many datasets that provide multi-class labels for object detection, LostAndFound is a unique dataset that annotates small obstacles caused by lost cargos down to the height of 5 centimeters \cite{pinggera2016lost}. In total, 42 classes of small obstacles are labeled such as crate, cardboard box, plastic bag, ball, and many others. In a similar vein, Bosch Small Traffic Lights provides videos that contain traffic lights with a width as small as 2 pixels, which are even difficult for human eyes to distinguish \cite{behrendt2017deep}. In total, 13 types of traffic lights including their shapes and colors are annotated.

Regarding weather conditions, many datasets contain scenes with rain, overcast, or even snow. Particularly, HCI Challenging Stereo provides 11 challenging sequences taken under extreme weather conditions which include scenes with flying snow, rain flares at night, snow at night, rain blurs, etc \cite{meister2012outdoor}.

\subsubsection{Behavioral and Contextual Data}
For the purpose of driver attention analysis, driver's gaze fixation is usually collected. While both DIPLECS and Elektra contain a moderate-sized data subset that provides the gaze information, the only large-scale dataset with accurately measured gaze fixation is Dr(eye)ve. In Dr(eye)ve, the gaze data is projected onto the video captured by a front camera, which can be used to predict driver's intent, improve road safety, and plan better driving strategies \cite{alletto2016dr}. Instead of gaze data, Brain4Cars recorded videos of drivers inside the car. Together with the videos of the road and the labels of six classes of vehicle maneuvers such as stop, turn and lane change, Brain4Cars is used for anticipating maneuvers several seconds in advance \cite{jain2015car}. Another dataset that focuses on driver's behavior is UAH, which simulates normal, drowsy and aggressive driving and records 7 maneuvers such as lane-drifting, overspeeding, car-following, etc.

While the above datasets only focus on driver behavior analysis, JAAD labels the behaviors of both drivers and pedestrians occurring in the same scene. Driver behaviors include stopping, moving slow, accelerating, etc, whereas pedestrian behaviors include actions such as crossing, looking, slowing down as well as their moving directions and whether they are at an intersection \cite{kotseruba2016joint}. The JAAD dataset is very useful for studying joint attention of drivers and pedestrians, which is still a challenging problem that awaits effective solutions \cite{rasouli2018joint}, as mentioned in Section \ref{sec:challenges}.

\section{Filling the Data Gap}
\label{sec:learninggap}
While there is an increased emphasis on dataset acquisition, the available data is still insufficient for learning an autonomous driving model that can operate robustly anywhere at anytime. In this section, we will discuss different ways to fill this data gap. We address the following three questions: 1) in which areas do we need to collect more naturalistic driving datasets? 2) how can one utilize synthetic data to complement real data? and 3) how can one use driving simulators for developing and testing learning algorithms? We conduct a brief discussion on current research efforts, open issues, and possible future directions related to these topics. 

\subsection{Targeted Data Acquisition}
The robustness of autonomous driving models depends on continuously resolving real-world hazardous corner cases, which requires continuously collecting datasets that expose the true diversity of the driving environment globally and include rich scene representations. In the following, we list several directions for new dataset collection that complement existing datasets.
\begin{itemize}
\item For joint attention studies and intent prediction, datasets that simultaneously record the behaviors of pedestrians, cyclists, and drivers of both the ego car and neighboring vehicles are needed. Behaviors include but are not limited to facial expressions, gestures and gaze. It would be better that such data is collected during an interactive event between multiple road users, e.g., a pedestrian establishes eye contact with the driver or the car before crossing the street, because many accidents can be avoided by interactions using eye contact or simple gestures \cite{rasouli2018joint}.

\item For robust obstacle detection, more datasets that capture uncommon hazards are needed, such as small pieces of cargo or debris falling from a vehicle in front of the ego car, accident scenes, road construction, etc. Recently, a Near-miss Incident DataBase was introduced in \cite{takahashi2016explaining}, but it is now under reconstruction and charges a fee for access. 

\item For deep scene understanding, datasets with multi-level annotations are helpful which include low-level object annotation, mid-level trajectory annotation and high-level behavior and relationship annotation \cite{xue2018survey}.

\item For driveability evaluation of various types of scenes, contextual scene descriptions are needed. The data collected is better tagged according to driving contexts such as intersections/roundabouts/non-intersections, signaled/unsignaled pedestrian crossing, number of marked/unmarked lanes, and so on, which could facilitate adaptive driving policy learning in various driving contexts. 
\end{itemize} 

Though not discussed here, the data collected could have dependency on the hardware and sensor calibration used during data acquisition. Many datasets provide calibration files for users to get exact positions of sensors and better use the measurements. Nonetheless, the goal is to learn an autonomous driving model that is invariant to hardware changes and work across different platforms. 

\subsection{Synthesizing Data}
It can be costly and cumbersome to collect data in the physical world and annotate them at a large scale. Therefore, researchers also resort to synthetic datasets that contain visually realistic images and automatically generated annotations for autonomous driving studies. Generating synthetic data is considered to be one major data augmentation approach for learning-based algorithms \cite{wong2016understanding}. Synthesizing data can create additional samples capturing conditions not covered in the real-world datasets, and can be used to augment both training and testing sets. Some well established synthetic driving datasets include Virtual KITTI \cite{gaidon2016virtual}, Synthia \cite{ros2016synthia}, VIPER \cite{richter2017playing}, etc, which are generally produced by video game engines. 

The biggest concern with synthetic data is that it is dependent on the data generation model and therefore may present bias and not generalize well enough to the physical world. Fortunately, thanks to the continuous development of game engines and learning algorithms such as generative adversarial networks (GANs) \cite{goodfellow2014generative}, the realism of synthetic images keeps increasing, which can then better generalize the models trained on these synthetic data to real-world scenarios. For example, it has been shown in \cite{shrivastava2017learning} that a CNN-based hand pose estimator trained only on synthetic data generated by GAN can even outperform a CNN trained on real images when applied to a realistic test dataset. Similar approaches are worth pursuing for training autonomous driving models. 

Besides training, synthetic data can also be used to test and identify the limitations of the trained model by simulating not yet encountered conditions. To this end, two very recent studies \cite{tian2017deeptest, zhang2018deeproad} focus on DNN testing by generating realworld-like images that cover scenes with extreme weather conditions to detect erroneous behaviors of a trained deep end-to-end learning model, where \cite{tian2017deeptest} uses affine image transformations and \cite{zhang2018deeproad} uses GAN to generate the test images. The underlying assumption is that at the same location, the predicted car maneuver should be similar under different weather conditions or illuminations. Both approaches have been reported to detect thousands of erroneous behaviors of top-ranked DNN models in the Udacity challenge \cite{Udacity}. While these findings are enlightening, the coverage of the generated test images in both studies are still limited to only rain and snow conditions. Extended synthetic datasets are needed to test DNNs in additional conditions and in more complex scenarios. Further robust mechanisms to assess and verify generalizability to real-world are needed.

\subsection{Driving Simulators}
As pointed out in \cite{schoner2018simulation}, it would need test driving for hundreds of millions of miles without accident to prove that an autonomous driving system is safe enough to be adopted. This is in general impractical and still cannot cover all conceivable scenarios. Therefore, simulators play an important role in simulating different scenarios and performing exhaustive tests for both ADAS and autonomous cars. Many simulators have been developed in recent years, such as CARLA \cite{dosovitskiy2017carla}, Microsoft's AirSim \cite{shah2018airsim}, Baidu's Apollo \cite{apollosimulation}, and NVIDIA's DRIVE Constellation \cite{nvidiadrive}, just to name a few. These simulators generate driving scenarios from user-defined models of roads, traffic and vehicles, which allow testing of autonomous driving models and discovery of interrelationship among various driveability factors. However, even though high-fidelity environments are simulated, the conditions there are generally fully controlled and can be different from real-world situations that have higher uncertainty and dynamics. Therefore, one should be aware that the conclusions drawn from using these simulators may not be easily translated to real-world scenarios. 

With various data sources of both realistic and simulated data, a collective approach is needed to integrate them and reap the most benefit from all available data sources. To this end, the PEGASUS research project \cite{PEGASUS} develops a data processing pipeline that stores data from different sources such as field tests, driving simulator studies, and traffic simulations into a database \cite{putz2017system}. The main objective of constructing this database is to collect relevant traffic scenarios and establish a common evaluation basis for autonomous vehicle testing and validation.

\section{Further Discussion}
\label{sec:discussions}
While this paper focuses on supervised learning that requires sufficient data, we would like to point out that additional advancements are needed in artificial intelligence (AI) to learn from limited data and generalize to unknown environments. Ideally, the AI engine that powers autonomous driving should have the capability of transferring models trained on source data to any target domain. \color{black} For example, if a driving model is trained under good lighting condition during midday, will it have similar performance at all times during the day, at different geographic locations, and under all weather conditions? Building a robust model that performs well in all those conditions may require extension or augmentation through unsupervised learning and reinforcement learning techniques that can transfer knowledge and driving policy to unseen scenarios. 

Many reinforcement learning based approaches have been reported for driving policy adaptation. For example, a RNN-based approach is proposed in \cite{ebrahimi2017gradient} which first learns an optimal initial policy architecture from expert demonstration, and then adapts this policy to a new driving domain using the rewards obtained in the new domain. To better understand a new driving context, it is also important to find a representation of a scene such that the difference between the source and the target scenes and the error made by learning models can be minimized \cite{pan2010survey}. To address this issue, \cite{japuria2017casnsc} adds semantic clues from the environment such as the distance of a pedestrian from the curbside and traffic light status to make pedestrian motion prediction more robust and flexible in new environments. Overall, the questions of what to transfer, when to transfer and how to transfer \cite{pan2010survey} need to be understood better in the context of autonomous driving. 

One last remark we would like to make here is that robust AI is only one component contributing to safe autonomous driving. Achieving a safe autonomous vehicle requires solving interdisciplinary problems in domains of computing hardware, robotics, security, social acceptance and many others \cite{7823109}, that are beyond the scope of this paper.


\section{Concluding Remarks}
\label{sec:conclusions}
In this paper, we have reviewed recent research efforts from a driveability perspective for autonomous driving. We presented both explicit and implicit factors that contribute to scene driveability, identified the potential risks posed by these factors, and investigated existing methods and metrics used for driveability assessment. With these investigations, we have shown the necessity of pursuing principled metrics for driveability, which can be represented by either a set of novel sophisticated metrics or a composition of metrics. It should also be well understood how the driveability metric interacts with other metrics used in system level verification and validation, which will have implication on optimization of multiple metrics concurrently and the trade-off therein. 

Furthermore, we have conducted an exhaustive overview of 45 open datasets collected on public roads and categorized these datasets according to use cases. More importantly, we highlighted datasets that are more suitable for training robust driving models and identified the scenarios that need more data acquisition. We have also proposed ways to fill the data gap including conducting targeted dataset acquisition, using synthetic data for training and testing, exploring driving simulators, and transferring knowledge to unseen scenarios. We hope this paper can serve both as a guidance on dataset selection and construction and as an invitation to pursue novel approaches that enable autonomous cars to navigate through all environments safely and reliably. 


\section*{Acknowledgement}
The authors would like to thank Prof. Maxim Likhachev from Carnegie Mellon University for his invaluable comments that improved the manuscript.

\bibliographystyle{IEEEtran}
\bibliography{drivability}

\end{document}